\documentclass{article}
\usepackage{acra}
\usepackage{graphics} 
\graphicspath{{figures/}}
\usepackage{epsfig} 
\usepackage{mathptmx} 
\usepackage{times} 
\usepackage{amsmath} 
\usepackage{amssymb}  
\usepackage{caption}
\usepackage{subcaption}
\usepackage{lipsum}

\usepackage[maxfloats=256]{morefloats}

\title{Radio Source Localization using Sparse Signal Measurements from Uncrewed Ground Vehicles}
\author{Asanka Perera $^{1,2}$, Vu Phi Tran $^{1}$, Sreenatha Anavatti $^{1}$, Kathryn Kasmarik $^{1}$, Matthew A. Garratt $^{1}$ \\ 
$^{1}$ University of New South Wales, Canberra, Australia \\ 
$^{2}$ University of Southern Queensland, Brisbane, Australia \\ 
asanka.perera@unisq.edu.au, \{phi.tran, s.anavatti, k.kasmarik, and m.garratt\}@adfa.edu.au}

\begin{document}

\maketitle

\begin{abstract}
Radio source localization can benefit many fields, including wireless communications, radar, radio astronomy, wireless sensor networks, positioning systems, and surveillance systems. However, accurately estimating the position of a radio transmitter using a remote sensor is not an easy task, as many factors contribute to the highly dynamic behavior of radio signals. In this study, we investigate techniques to use a mobile robot to explore an outdoor area and localize the radio source using sparse Received Signal Strength Indicator (RSSI) measurements. We propose a novel radio source localization method with fast turnaround times and reduced complexity compared to the state-of-the-art. Our technique uses RSSI measurements collected while the robot completed a sparse trajectory using a coverage path planning map. The mean RSSI within each grid cell was used to find the most likely cell containing the source. Three techniques were analyzed with the data from eight field tests using a mobile robot. The proposed method can localize a gas source in a basketball field with a 1.2 m accuracy and within three minutes of convergence time, whereas the state-of-the-art active sensing technique took more than 30 minutes to reach a source estimation accuracy below 1 m.
\end{abstract}

\section{INTRODUCTION}

Mapping the distribution of gas is a critical task for environmental monitoring, disaster response, and safety inspections. This process enables accurate evaluations of gas concentrations in various scenarios. Conducting outdoor experiments with gases presents challenges, prompting the consideration of radio sources as a viable alternative to conventional gas sources. Radio source localization refers to the estimation of the position of a radio signal source, such as a transmitter or a reflection of a signal, in a three-dimensional space~\cite{sun2020hidden,frew2005radio}. In wireless communication, radio source localization is an important aspect of various applications, including wireless sensor networks, positioning systems, and surveillance systems. Locating the true radio source position is a challenging problem due to the various factors involved in radio transmission. The transmitter, receiver and environment all play vital roles in achieving reliable communication. The most common reasons for highly dynamic radio signal reception are free space losses, multi-path reflections, obstacles, refraction, diffraction, and scattering \cite{sizun2005radio}. These factors can contribute to fluctuating, noisy or weak signal reception.

In recent years, particle filters have emerged as a powerful tool for solving this problem~\cite{wagle2010particle,charrow2014cooperative,xu2019locating}. Particle filters are a class of Bayesian filters that use a set of particles to approximate the posterior distribution of the state of a system. In radio source localization, the state of the system is the location of the source, and the observations are the received signal strengths at a set of receivers. In our recent work~\cite{tran2023multi}, we investigated an improved particle filter-based gas source localization approach using multiple mobile ground robots. The basic idea behind particle filters for radio source localization is to use a set of particles to represent the possible locations of the source. Each particle is a hypothesis about the location of the source, and its weight is determined by how well it explains the received signal strengths. There are many variations of particle filters that have been used for radio source localization, including sequential Monte Carlo methods~\cite{abu2017sequential,bandiera2015cognitive}, unscented particle filters~\cite{qian2023direct}, and Rao-Blackwellized particle filters~\cite{jung2014indoor}. A main limitation of particle filters is their computational complexity, especially when the number of particles is large.

Active sensing is considered as the state-of-the-art in radio or gas source localization~\cite{TRAN2022101171,tran2023coverage}. Our simulated and empirical studies indicated that it takes a relatively long period of time to complete radio source localization. For example, a particle filter coupled standard active sensing method took roughly 32 minutes to collect 160 measurements from a $3\times6$ m area with three Jackal robots, achieving a mean square error (MSE) of 0.68~\cite{tran2023multi}. In contrast, the proposed method took roughly 8 minutes to collect measurements from an area of size $15\times28$ m (a basketball field) using only one Jackal robot to estimate the source location with an error of 1 m. Although the experimental parameters were different, we can clearly draw a conclusion about the quick turnaround and processing time of the proposed technique. 
 While particle filters can achieve high accuracy in the estimation with sufficient time and computational cost, there is a need for alternative methods of radio source localization that achieve reduced computational complexity and shorter turnaround time.

We address these challenges by leveraging a coverage path planning algorithm~\cite{TRAN2022101171} to collect sparse readings. This approach eliminates the robot backtracking while traversing a map with a set time budget. The data was processed by a novel source estimation technique with reduced computational complexity. A total of three methods were analyzed for the source estimation. The results from 8 data collections show that the proposed method can localise the radio source with high accuracy.

The contributions of this work are:
\begin{itemize}
    \item 
    A new radio source localization technique is introduced to achieve a faster turnaround time with a low computational complexity. The Received Signal Strength Indicator (RSSI) measurements were collected while the robot completed a sparse trajectory using a coverage path planning map. The proposed approach uses the mean RSSI within each grid cell to find the most likely cell containing the source.
    \item The radio propagation model and the Fresnel zone characteristics are analyzed for our experimental setup.
    \item A comparative study was performed, benchmarking two interpolation algorithms.
    \item Extensive outdoor experiments were conducted to analyze the radio signal propagation between the robots and validate the performance of the proposed technique.
\end{itemize}

The remainder of this paper is organised as follows: Section II discusses the works related to the proposed study, Section III presents our methodology, and Section IV presents our experiments, results and discussion. We conclude in Section V.

\section{RELATED WORKS}

Radio source localization using uncrewed ground robots and uncrewed aerial vehicles (UAVs) has been an active research area in recent years. The location of the radio source can be estimated by the measurements collected by single or multiple receiver antennas. Here, we briefly discuss some notable works related to our study.

Different robotic platforms have been adopted for radio source localization. Single robot~\cite{huo2020autonomous,twigg2012rss,song2012simultaneous}, multi-robot~\cite{charrow2014cooperative,zhang2021assembling}, and UAV-based~\cite{isaacs2014quadrotor,hasanzade2018rf} approaches are prominent in robot-assisted radio source localization studies.

The mobility of the signal measuring platform has been combined with various source-finding algorithms. Twigg et al.\cite{twigg2012rss} used an RSSI gradient finding technique to efficiently explore a complex indoor environment while seeking a radio source. The algorithm estimates the 2-D RSSI gradient to prioritize candidate frontiers and avoid random or exhaustive exploration. The aim of the study was to move the robot towards the radio source, hence limiting the robot's ability to explore for multiple radio sources. Moreover, the often fluctuating RSSI readings can make the gradient-based technique unreliable. In a similar study, a single mobile robot was used to localize multiple unknown transient radio sources using their posterior probability distributions\cite{song2012simultaneous}. A ridge-walking motion planning algorithm was proposed to enable the robot to efficiently traverse high-probability regions and accelerate the convergence of posterior probability distributions.

A team of robots can provide better information and faster coverage of the field, as demonstrated by the following works. In one study, a team of robots was employed to locate an unknown target in a complex, known environment using range sensors~\cite{charrow2014cooperative}. The robots were equipped with radio-based time-of-flight range sensors and adopted a Bayesian approach for estimation. The researchers presented a control law that maximizes the mutual information between the robot's measurements and their current belief of the target position. Practical applications of multiple robots for radio source localization in Mount Etna, Sicily, Italy, have been demonstrated in~\cite{zhang2021assembling,staudinger2023enabling}. They used a phased array antenna to estimate the signal direction of the arrival of a radio beacon with an unknown position. The position estimate was performed using a decentralized particle filter method. The RSSI readings collected by the robots can be unreliable due to external factors at times. To cope with this situation, an information-sharing mechanism was introduced in~\cite{kim2011localization}. The robots paired up and used signal strength ratios to obtain the sensed conditional joint posterior probability of source locations for the team, which can be derived from pairwise joint posterior probabilities.

The increased mobility of UAVs can help in recording less noisy signal power measurements in challenging geographies. For instance, Isaacs et al. (2014) adopted a particle filter algorithm for UAV-based source localization and tracking in complex indoor environments with multipath fading effects. Similarly, Hasanzade et al. (2018) proposed a solution for localizing an RF source over a large-scale environment using UAVs and RSSI. They evaluated the noise effect on RSSI through a measurement test and used a particle filter for the localization process. However, both studies were limited to indoor or simulation environments.

The use of various types of antennas has been the primary hardware arrangement for many research studies. Single or multiple directional antennas are the widely used antenna type for radio source localization~\cite{isaacs2014quadrotor,song2012simultaneous,zhang2021assembling,hasanzade2018rf}. They provide the angle of arrival information of the received signal. In our study, we did not equip our robot with directional antennas. Instead, whip antennas were used to broaden the applicability of our method to other sources with similar distribution characteristics such as gas or radioactive sources where it is difficult or impossible to measure direction. RSSI is a widely used metric radio signal propagation. It can be used for a variety of applications, such as monitoring signal strength, estimating the distance between the transmitter and the receiver, and performing signal strength-based location estimation.

All the above works use RSSI measurements. Most of the studies use particle filter-based techniques to locate the radio source, which can require a higher number of particle updates to build a reliable probability map. This, in turn, requires the robot to travel to random or overlapping regions. Our proposed technique aims to address this challenge by estimating the source location from RSSI readings obtained from a sparse robot trajectory.

\section{METHODOLOGY}
This section explains the techniques employed to move a mobile robot in an outdoor environment to sweep an area for radio signal reception, the radio signal propagation model, practical issues of implementing them in robots, and radio source localization from sparse RSSI measurements

\subsection{Signal propagation model}
In this sub-section, we discuss the Friis path loss equation, which was used as a reference for error performance analysis of RSSI distribution plots. The Friis free space propagation model~\cite{friis1946note} is commonly used to model line-of-sight (LOS) path loss in a free space environment. This model assumes that there are no objects present that may lead to absorption, diffraction, reflection, or any other phenomenon that could alter the characteristics of a radiated wave. The propagation path loss between two isotropic antennas in free space, denoted by $P_L$, can be calculated using equation \ref{eq:PL} below:

\begin{equation}\label{eq:PL}
P_L(dB) = -10 log_{10} \frac{\lambda^2}{(4 \pi d)^2}
\end{equation}

where $\lambda$ is the signal wavelength and $d$ is the distance between the two antennas.

We used XBee Wi-Fi modems (IEEE 802.11n standard) operating at a frequency of 2.4 GHz for both the transmitter and receiver. The Friis equation was modified according to the specifications provided in the XBee data sheet. Each antenna can be unique and need to account for specified power ratings by the manufacturer.  The XBee has a sensitivity of -71 dBm and the transmitter was set to the lowest power level of 0 dBm (1 mW), even though it has five power levels (0 dBm to 4 dBm). The transmitter has an additional transmit power of 13 dBm. The free space path loss exponent is chosen on the basis of how cluttered the environment is. We selected a value of $n$ appropriate to our test environment using the parameters suggested in~\cite{barclay2003propagation}. The received power $P_r$ at a distance $d$ from the transmitter can be calculated using the Friis equation, which we have modified to include our system parameters.
\begin{equation}\label{eq:FZ}
P_r(dBm) = 13 (dBm) + 20 log_{10} \frac{\lambda}{(4 \pi)} - 10 n* log_{10} d + 71 (dBm)
\end{equation}

\begin{figure}
\centering
\includegraphics[width=0.35\textwidth]{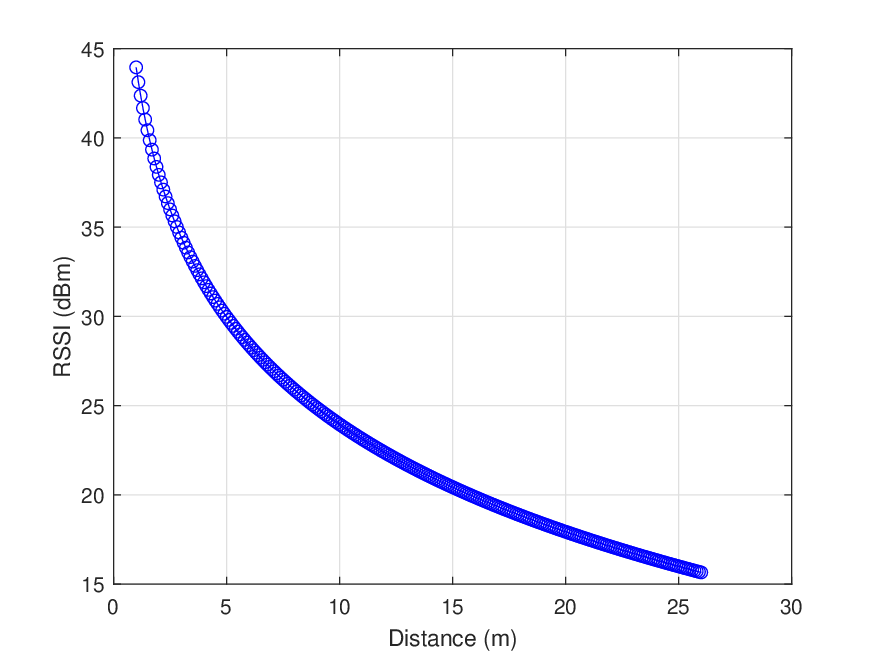}
\caption{The Friis path loss equation was used to calculate a path loss model. This model was used as a reference for error performance analysis of RSSI distribution plots.}
\label{fig:friis}
\end{figure}

The Friis equation is illustrated in Fig.~\ref{fig:friis}. When calculating the propagation model, we did not consider the antenna gains.

In our experimental setting, the XBee modems were mounted 0.6 m above ground level. This setup can cause significant signal distortions due to reflections from the close ground plane and obstruction of the Fresnel zone. In the following sub-section, we will discuss this issue in detail.

\subsection{Fresnel zone effect on the radios mounted on robots}

\begin{figure}
     \centering
     \begin{subfigure}[b]{0.25\textwidth}
         \centering
         \includegraphics[width=\textwidth]{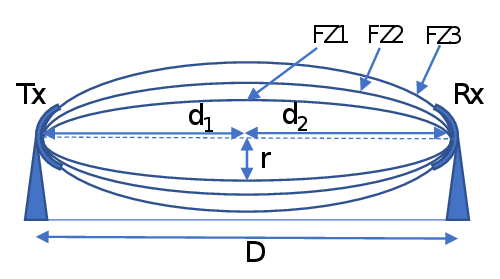}
         \caption{The first three Fresnel zones between the transmitter and receiver.}
         \label{fig:fz_th_1}
     \end{subfigure}
     \hfill
     \begin{subfigure}[b]{0.35\textwidth}
         \centering
         \includegraphics[width=\textwidth]{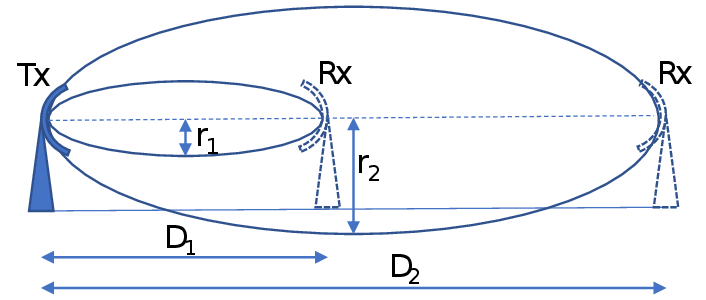}
         \caption{Change of Fresnel zones for different spacing between the transmitter and receiver. When the Fresnel zone expands the ground plane becomes an obstacle.}
         \label{fig:fz_th_2}
     \end{subfigure}
        \caption{Here, FZ stands for the Fresnel zone and Tx and Rx represent the transmitter and receiver respectively.}
        \label{fig:fz_th}
\end{figure}

The Fresnel zone represents a 3D elliptical region between the transmitter and receiver antennas where the signal strength is the strongest~\cite{sizun2005radio}. The size of this region depends on the distance between the antennas and the signal frequency. When designing radio transmission links, the first three Fresnel zones are typically taken into account. Obstructions within the Fresnel zone can result in signal loss at the receiver, even if the obstacles are not in the line of sight path. For reliable communication, it is recommended that at least $80\%$ of the first Fresnel zone be obstruction-free~\cite{coleman2012cwna}. However, the maximum recommended obstruction is $40\%$, which leaves only $60\%$ of the area unobstructed for transmission. While Fresnel zones 2 and 3 are also important, the first zone is the most critical as it covers the region of highest signal strength.

As demonstrated in Fig.~\ref{fig:fz_th_1}, consider a point at a distance $d_1$ from the transmitter and $d_2$ from the receiver on the line of sight path. The Fresnel zone radius at that point $r_n$ can be calculated using equation \ref{eq:fz}.

\begin{equation}\label{eq:fz}
r_n = \sqrt{\frac{n d_1 d_2 \lambda}{d_1 + d_2}}
\end{equation}
Here, $n=\{1,2,3\}$ is the Fresnel zone number.

Assuming both antennas have equal heights $h$, the maximum obstruction from the ground plane occurs when the Fresnel zone radius is at its maximum. The corresponding point lies at the center of the line-of-sight path, where $d_1 = d_2 = D/2$. The wavelength can be calculated using $\lambda = c/f$, where $c$ is the speed of light in air and $f$ is the transmission signal frequency. At the midpoint of the line-of-sight path, equation~\ref{eq:fz} can be rewritten as follows.

\begin{equation}
r_n = \sqrt{\frac{n D c}{4 f}}
\end{equation}

For the transmitter and receiver, we used XBee WiFi antennas. More details of the hardware setup are given in Section~\ref{sec:experiments}. The antenna height of our test platform was $0.6$ m. At this low height, the ground plane becomes an obstacle for relatively long-distance transmissions. This scenario is illustrated in Fig.~\ref{fig:fz_th_2} for two different distances between the transmitter and receiver. At $D_2$, the Fresnel zone is obstructed by the ground plane. We analyzed the ground plane obstruction for signal loss as follows.

When $r<h$, the obstruction-free Fresnel zone percentage $FZ_p$ is $100\%$, and it decreases according to the following equation when $r>h$.

\begin{equation}
FZ_p = 100 h/r_n
\end{equation}

\begin{figure}
\centering
\includegraphics[width=0.4\textwidth]{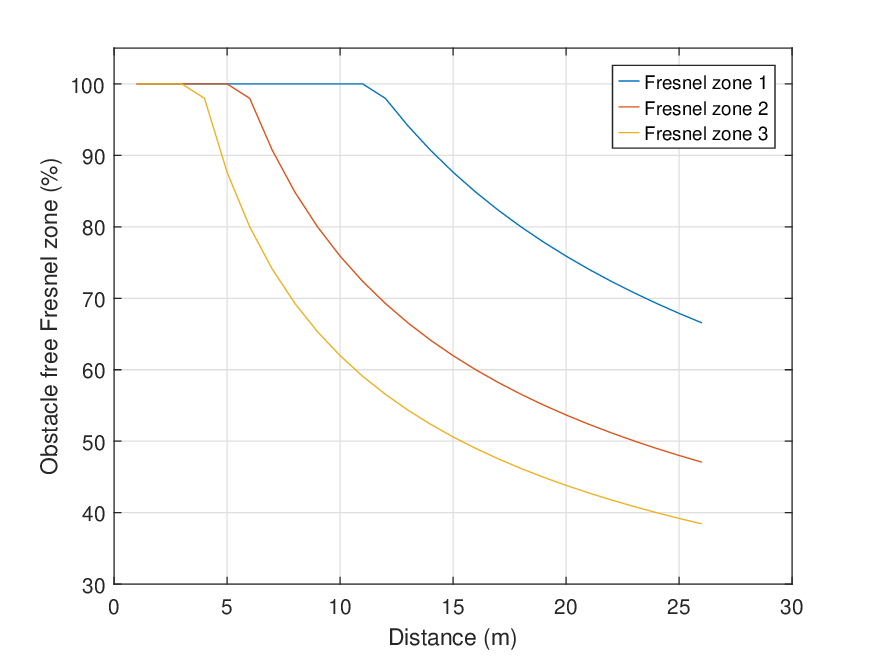}
\caption{The obstacle-free Fresnel zone variation with distance is shown. This modelling is for a transmitter and receiver with 0.6 m height each.}
\label{fig:fresnelzone}
\end{figure}

The variation of obstacle-free Fresnel zone percentage with distance is plotted in Fig.~\ref{fig:fresnelzone}. For the given antenna height, only the first Fresnel zone lies within the recommended $60\%$ obstruction-free region. That means both radios are able to communicate reliably within the tested $26m$ range. However, the third and second Fresnel zones start to get obstructed at $3m$ and $5m$ distances, respectively. The flat ground plane acts as a reflector for high-strength signals at these obstructed areas.

\subsection{Mobile robot coverage path planning}

We use a mobile robot coverage path planning algorithm to collect sparse radio strength measurements. For more comprehensive information regarding the coverage path planning algorithm, we suggest referring to our recent research study \cite{tran2023multi}. Initially designed for multi-robot application, we have repurposed the same algorithm for a single robot use case. Below, we provide an overview of the primary components of the algorithm.

\begin{itemize}
   \item In many scenarios, there are practical limitations on the allowable traversal time budget while also needing to maximize all information (mean and uncertainty). The algorithm incorporates a time budget to create the optimal robot path.

    \item We discretize the obstacle geometry on the map into a grid. A scanline algorithm \cite{clark1979} is utilized to differentiate the grid cells as either an obstacle or free space.

    \item A block-building algorithm is applied to group adjacent free grid cells into variable-sized blocks. The maximum block size is selected to accommodate the group of UGVs' (in multi-robot scenarios) spread-out size. The remaining block sizes are obtained by progressively halving the maximum size. After scanning the entire map to determine the suitable block sizes, we obtain a list of blocks covering the entire map. A minimum spanning tree algorithm is then used to find a spanning tree by removing connections that are not part of the spanning tree.

    \item A spanning tree is a subset of an undirected graph that connects all the graph vertices using the minimum number of edges possible. The cost of the spanning tree is the total weight of all edges in the tree. The minimum efficient spanning tree can be obtained using a minimum spanning tree (MST) algorithm. The MST algorithm constructs a tree containing every vertex, and the sum of the edge weights in the tree is minimized.

    \item The path planning is executed by dividing each block into four logical parts. The central point of the four parts is the connection point located on the constructed route.
\end{itemize}

\begin{figure}
\centering
\includegraphics[width=0.35\textwidth]{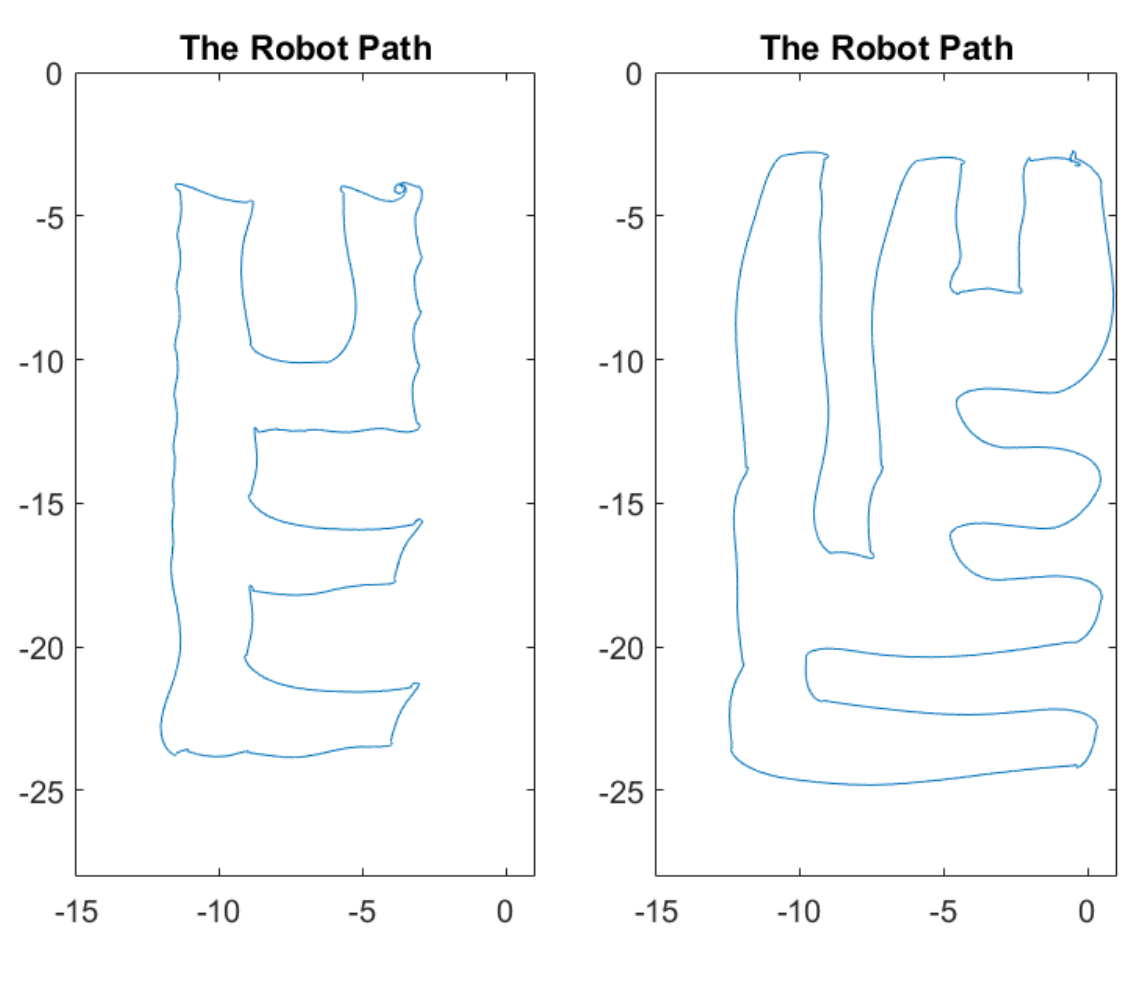}
\caption{Two maps were used for the experiments. The linear velocity of the robot was $0.8$ m/s and the angular velocity was $0.75$ m/s when traversing the shown maps. The above maps demonstrate the differential GPS coordinates recorded from the robot. The radio source was placed in the middle of the field at (-7.5, -14).}
\label{fig:maps}
\end{figure}

We utilized the aforementioned coverage path planning algorithm to generate two maps (refer to Fig.~\ref{fig:maps}) for our experiments. For each map, we adjusted only the time budget to create two varying cell sizes while retaining all other parameters constant.

\subsection{Proposed RSSI assignment method}

We implement a technique that averages the received power level to find the tile in the grid map with the highest probability of containing the radio source. Once the planned trajectory is complete, the source localisation calculation can be performed without any additional data gathering.

The entire map is divided into $a\times a$ tiles. We calculate the mean RSSI of the readings collected inside a particular tile area and assign it to that tile. However, since our robot traversed a sparse map, we were not able to generate mean RSSIs for the entire region of interest (a basketball court of $15\times28$ m). To test our approach, we initially divided the map area into $1\times1$ m tiles and calculated the mean RSSI for the tiles that the robot traversed. Fig.~\ref{fig:grids} shows a completed map with the RSSI assignments. In Fig.~\ref{fig:grids}.(b), only a fraction of the area is covered by RSSI measurements, and the rest of the area does not have corresponding RSSI readings. We consider the tile with the highest mean value to be the most likely location of the radio source.

\section{EXPERIMENTS AND RESULTS}\label{sec:experiments}

In this section, we discuss the hardware, ground truth data collection and analysis, and the available and proposed techniques for the radio source estimation.

\subsection{The experimental setup}

\begin{figure}
\centering
\includegraphics[width=0.25\textwidth]{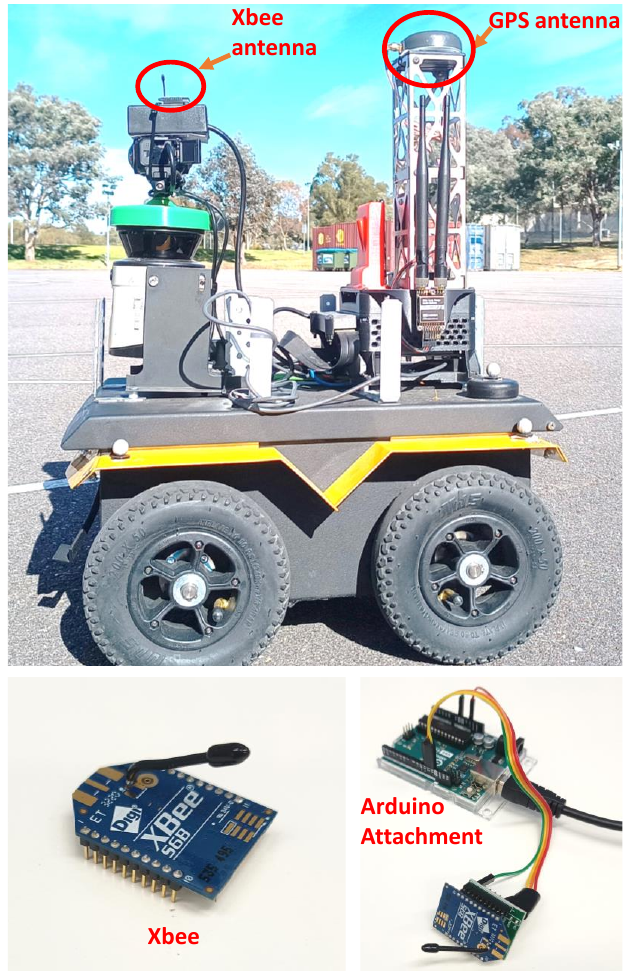}
\caption{The Jackal mobile robot used for the radio source localization field tests is shown on top. 
The XBee modems are mounted as high as possible on the robot to minimize ground reflections. Accurate positioning was achieved with a differential GPS module attached to the robot. The robot is also equipped with a LiDAR for obstacle avoidance. The XBee modem and the Arduino attachment used for the testing are also shown.}

\label{fig:hardware}
\end{figure}

We used a Jackal mobile robot (as shown in Fig.~\ref{fig:hardware}) as our moving platform for mounting the radio receiver. The Jackal robot is equipped with several sensors, including a differential GPS (DGPS) rover module, Inertial Measurement Unit (IMU), and a SICK LMS-111 LiDAR with an observation range of 4 meters. Incoming GPS rover signals and the internal IMU sensor were read at a rate of 10 Hz each. The DGPS base station was stationary and transmitted DGPS corrections to the rovers. The sampling time of the whole system was set at 30 Hz.

As our test area was relatively small (a basketball field without obstacles), we selected low-power XBee Wi-Fi modems (S6B model) as our transmitter and receiver. It supports RF data rates up to 27 Mbps. The XBee has a whip antenna that protrudes about 25 mm above the surface of the XBee PCB. One XBee was used as a transmitter on a fixed platform, and the other was mounted on the Jackal robot. The receiver XBee was connected to an Arduino computer on the robot platform, and the RSSI readings were transmitted to the robot through ROS (Robot Operating System) messages. The RSSI readings were collected at a frequency of 1 Hz and recorded for post-processing to localize the source.

\subsection{Ground truth data collection}

\begin{figure}
\centering
\includegraphics[angle=270, width=0.2\textwidth]{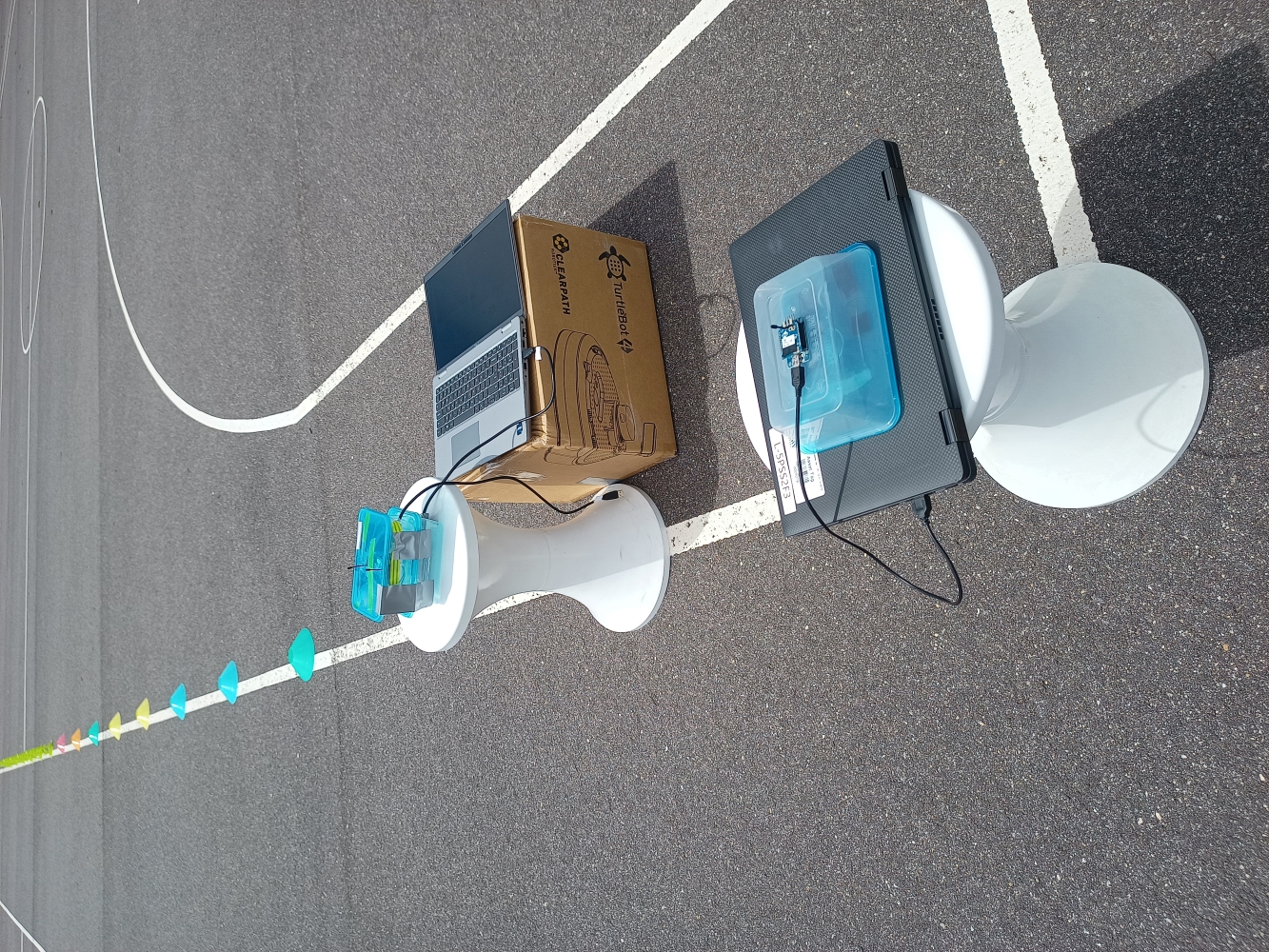}
\caption{An XBee receiver was placed at $1m$ intervals from the transmitter in an outdoor basketball field to collect RSSI ground truth data.}
\label{fig:fieldsite}
\end{figure}

\begin{figure}
     \centering
     \begin{subfigure}[b]{0.4\textwidth}
         \centering
         \includegraphics[width=\textwidth]{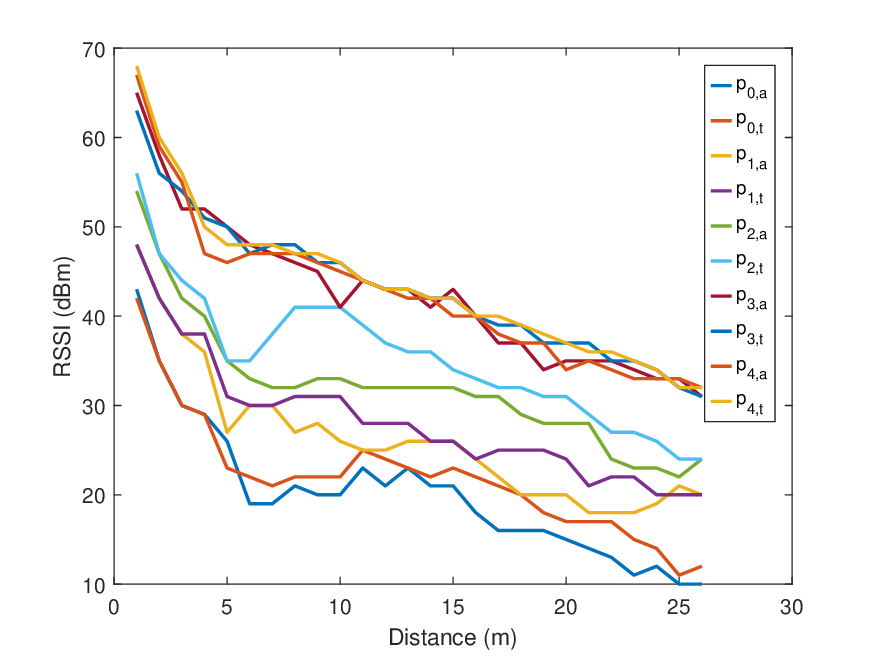}
         \caption{RSSI variation with different power levels and distances.}
         \label{fig:rssi_a_t}
     \end{subfigure}
     \hfill
     \begin{subfigure}[b]{0.4\textwidth}
         \centering
         \includegraphics[width=\textwidth]{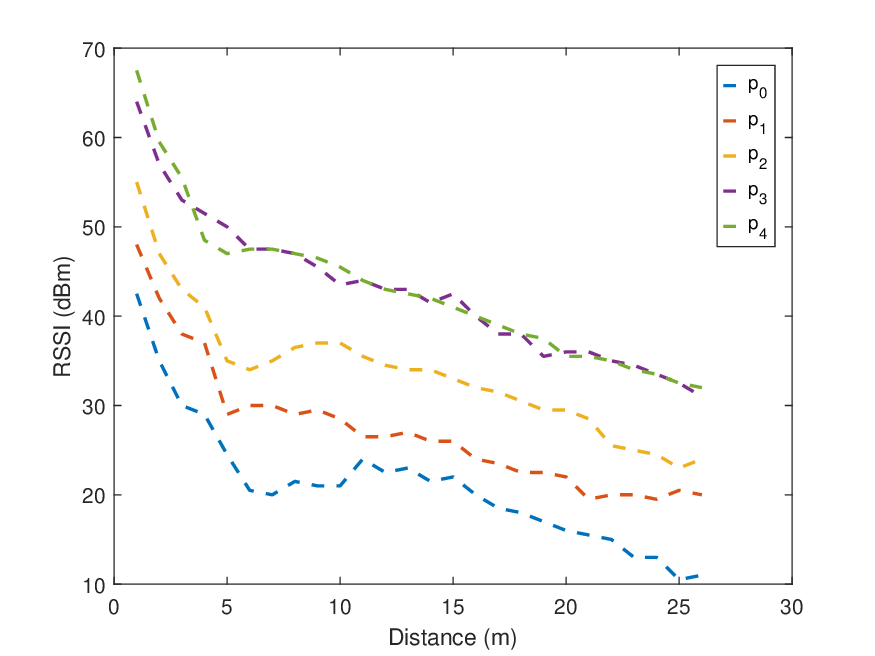}
         \caption{Average RSSI variation with different power levels and distances.}
         \label{fig:rssi_avg}
     \end{subfigure}
        \caption{RSSI readings were recorded at $1$ m intervals both when the receiver was moving away from and towards the transmitter. Here, $p_n$ represents the RSSI for a specific power level, with subscripts $a$ and $t$ denoting \textit{away from the source} and \textit{towards the source}, respectively, as $n$ indicates the power level from 0 dBm to 4 dBm with 0 being the weakest and 4 the strongest.}
        \label{fig:three graphs}
\end{figure}

When the robot is moving with the receiver, the received signal power can be affected by the movement and its sensors. Therefore, to analyze the signal power strength in the field site, we collected ground truth data with two XBees mounted on stationary platforms. As shown in Fig. \ref{fig:rssi_a_t}, the measurements were taken between 1m and 26m at intervals of 1m. The data collection was conducted while the receiver was moving away from and moving towards the transmitter under five different transmission power settings ($P_0$ = 0 dBm, $P_1$ = 5 dBm, $P_2$ = 10 dBm, $P_3$ = 15 dBm, and $P_4$ = 19.78 dBm). The analysis of this data collection is illustrated in Fig.~\ref{fig:rssi_avg}. All the RSSI graphs follow the overall path loss characteristics but the measurements are noisy.

The area of our field site was $15\times28$ m, which is relatively small compared to general RF transmission link areas. In order to have a strong signal near the source, minimize ground reflections, and obtain a relatively weak signal strength when the receiver is several meters away, we selected the minimum transmission power level of 0 dBm for further testing.


\subsection{RSSI data analysis}

\begin{figure}
\centering
\includegraphics[width=0.4\textwidth]{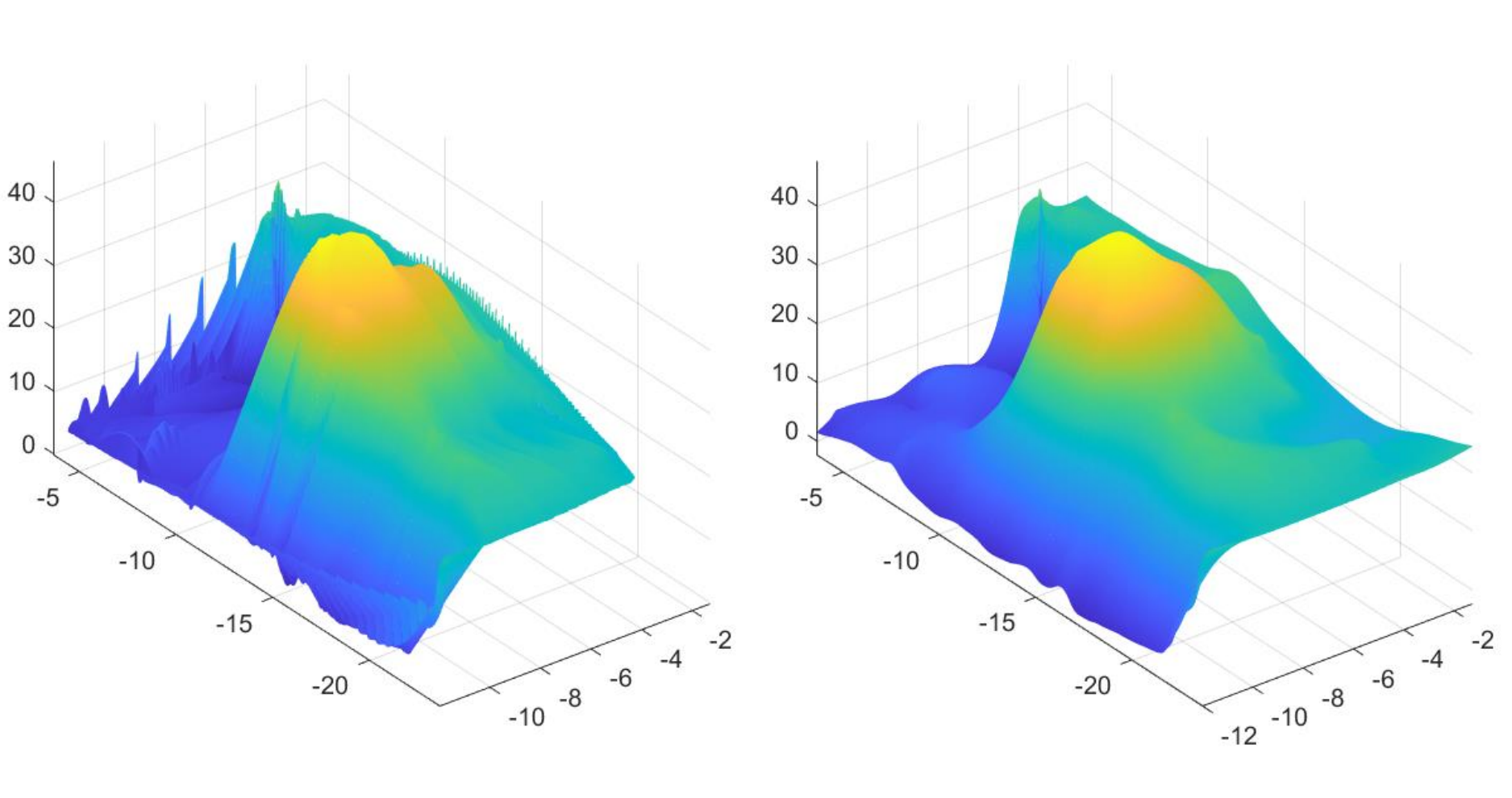}
\caption{The measured RSSIs were interpolated to a 3D surface map using Cubic (left) and Biharmonic spline (right) methods.}
\label{fig:surfplots}
\end{figure}

We use a low-time budget path to explore the area of interest. In this case, the signal power measurements are collected only along the robot trajectory. Therefore, the measurements are highly sparse. The collected data was post processed to estimate the radio source location. To estimate the missing surface data, interpolation techniques can be employed.

We experimented with different interpolation methods such as linear~\cite{watson2013contouring}, nearest neighbor~\cite{watson2013contouring}, natural~\cite{sibson1981brief}, cubic~\cite{watson2013contouring} and Biharmonic spline~\cite{sandwell1987biharmonic} to interpolate and build a 3D mesh of the received signal power. Among these methods, cubic and Biharmonic spline methods showed better RSSI maximum approximations compared to the other methods. Therefore, for the performance analysis, we include only cubic and Biharmonic spline (also referred to as V4) interpolation techniques. The interpolated 3D heat maps for one robot trial are shown in Fig.~\ref{fig:surfplots}. In both heat maps, the high signal strength near the source is clearly visible. However, the RSSI readings around the source are not consistent.

\subsection{RSSI assignment experiments}

\begin{figure}[ht!]
\centering
\includegraphics[width=0.495\textwidth]{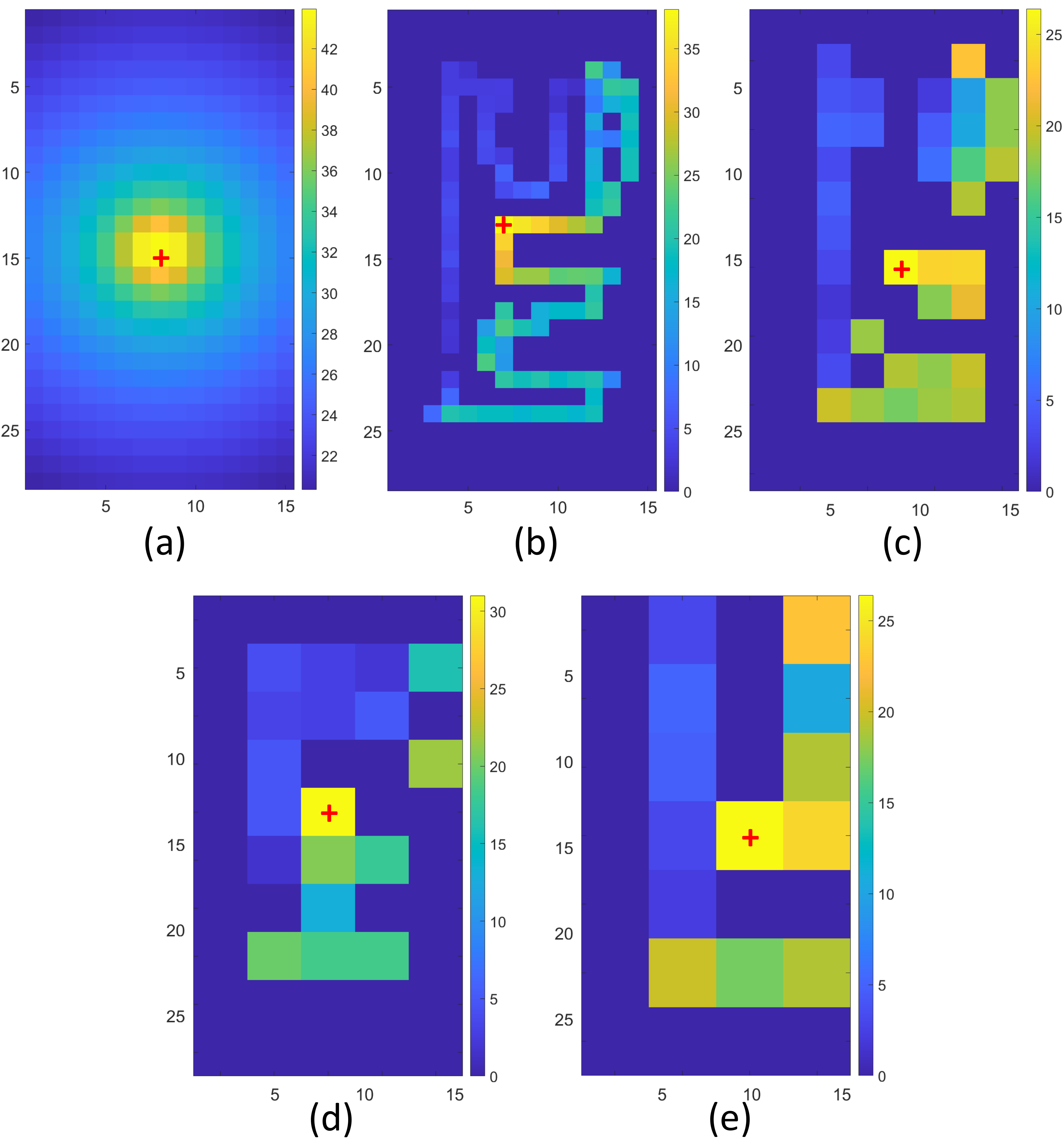}
\caption{(a) The simulated radio source using the Friis path loss equation is shown on the first image. Image (b) shows the RSSI readings assigned to $1\times1$ m tile. Images (c) to (e) show the calculated mean RSSIs for tile sizes of $2\times2$ m, $3\times3$ m, and $4\times4$ m respectively. Estimated source locations are shown as a red +.}
\label{fig:grids}
\end{figure}

We experimented with different tile sizes ($1\times1$ m, $2\times2$ m, $3\times3$ m, and $4\times4$ m) and analyzed their performance, as depicted in Fig.~\ref{fig:grids}. In the given example shown in Fig.~\ref{fig:grids}, Fig.~\ref{fig:grids}.(b) and Fig.~\ref{fig:grids}.(c) show the closest location estimate. Large tile sizes can be useful when dealing with high levels of uncertainty and noisy measurements.

The developed method helps identify the location of the radio source at different resolutions. Received power level readings often suffer from errors due to various conditions, such as reflection, scattering, and path losses. The ability to identify the most likely source location based on adjacent measurements is important. We select the most probable tile in the grid using sparse measurements and without the robot moving to new tiles. The center of that tile with the highest mean value gives the estimated source location. We consider this a simple yet effective method for localizing a highly dynamic radio wave origin.

\subsection{Source estimation comparison}

\begin{figure}[ht!]
\centering
\includegraphics[width=0.4\textwidth]{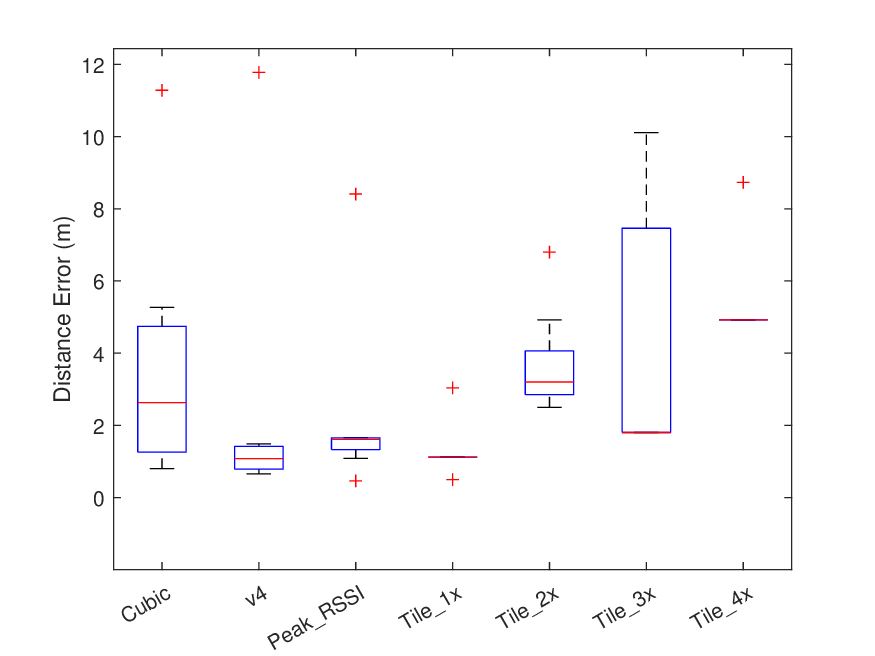}
\caption{The performance of different source finding techniques are compared against each other for their performance. Each error indicates the distance between the true source location and the estimated source location. Each column represents the error calculations from eight robot tests involving two different maps. \textit{Cubic} and \textit{V4} are the interpolation techniques. \textit{Peak\_RSSI} is the distance between source and the peak RSSI value. \textit{Tile\_1x} is the error when $1\times1$ m tiles are used for RSSI assignments. \textit{2x}, \textit{3x}, and \textit{4x} represent $2\times2$ m, $3\times3$ m, and $4\times4$ m tiles respectively.}
\label{fig:boxplot}
\end{figure}

\begin{figure}
     \centering
     \begin{subfigure}[b]{0.45\textwidth}
         \centering
         \includegraphics[width=\textwidth]{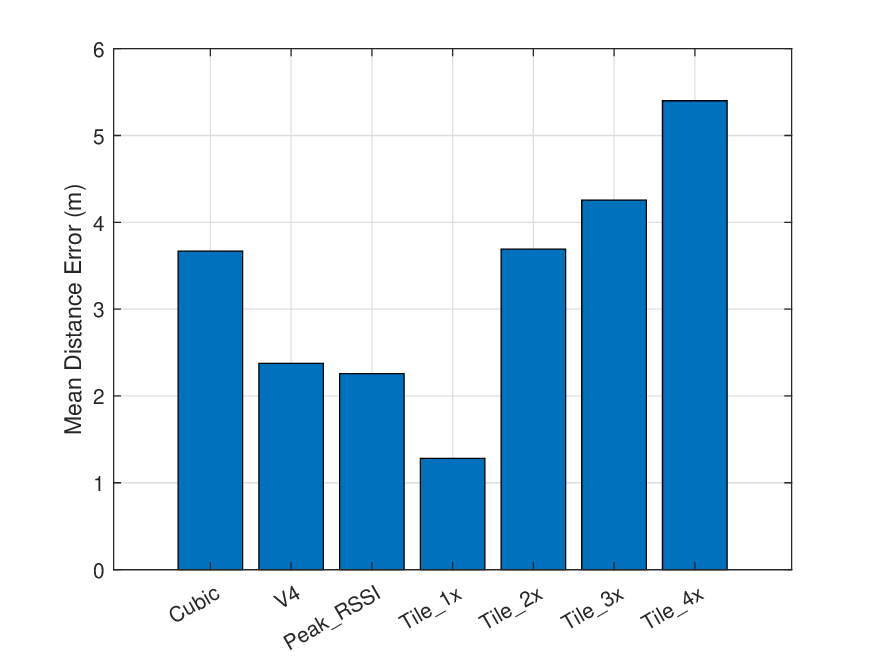}
         \caption{Mean distance error.}
         \label{fig:mean_dist}
     \end{subfigure}
     \hfill
         \begin{subfigure}[b]{0.53\textwidth}
         \centering
         \includegraphics[width=\textwidth]{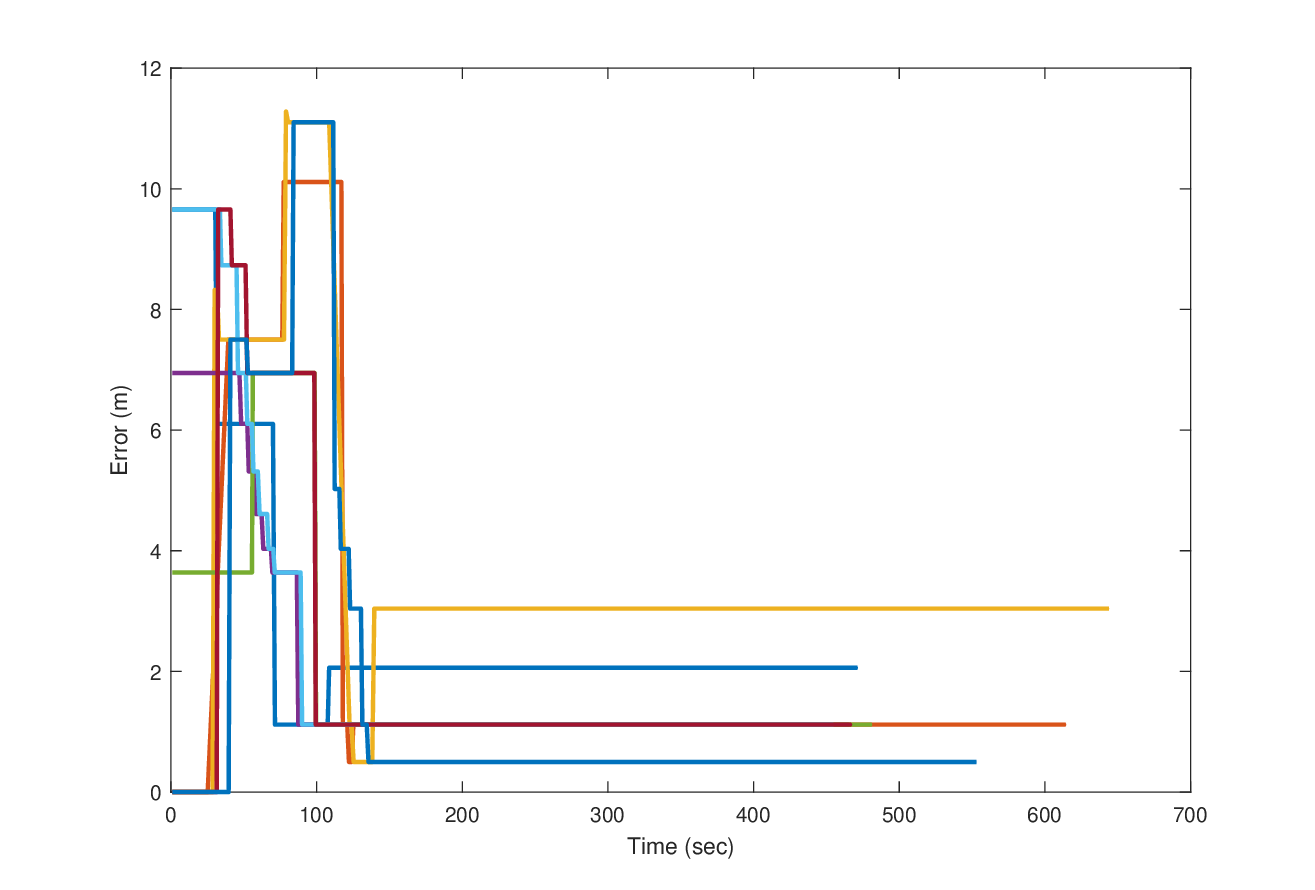}
         \caption{Error convergence of \textit{Tile\_1x} method (above).}
         \label{fig:error_time1}
     \end{subfigure}
     \caption{Mean distance error and error convergence graphs for 8 tests.}
     \label{fig:dist_time}
\end{figure}

\begin{figure}[]
\centering
\includegraphics[width=0.5\textwidth]{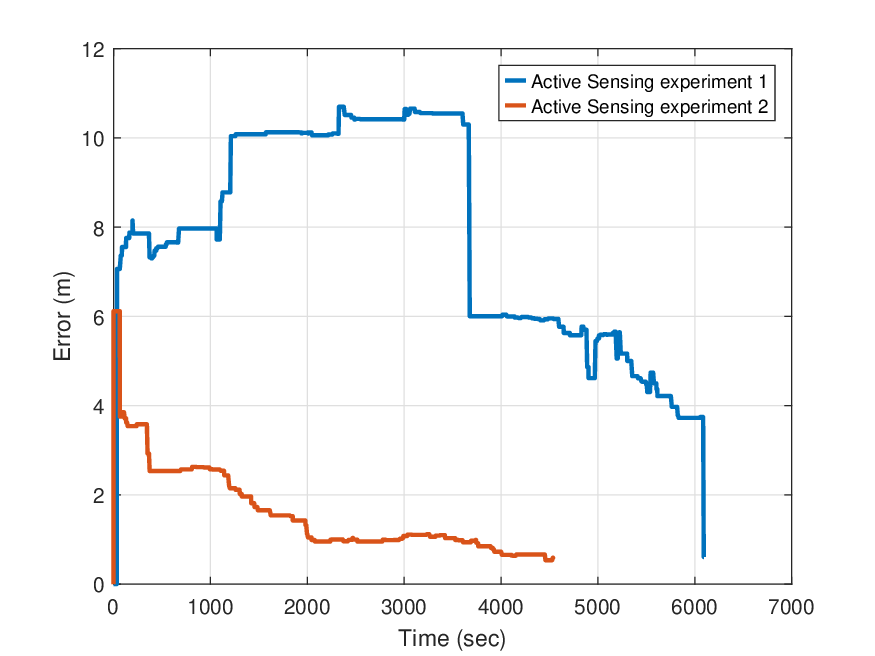}
\caption{Error convergence of active sensing experiments.}
\label{fig:as_err}
\end{figure}

\begin{figure*}[ht!]
\centering
\includegraphics[width=1\textwidth]{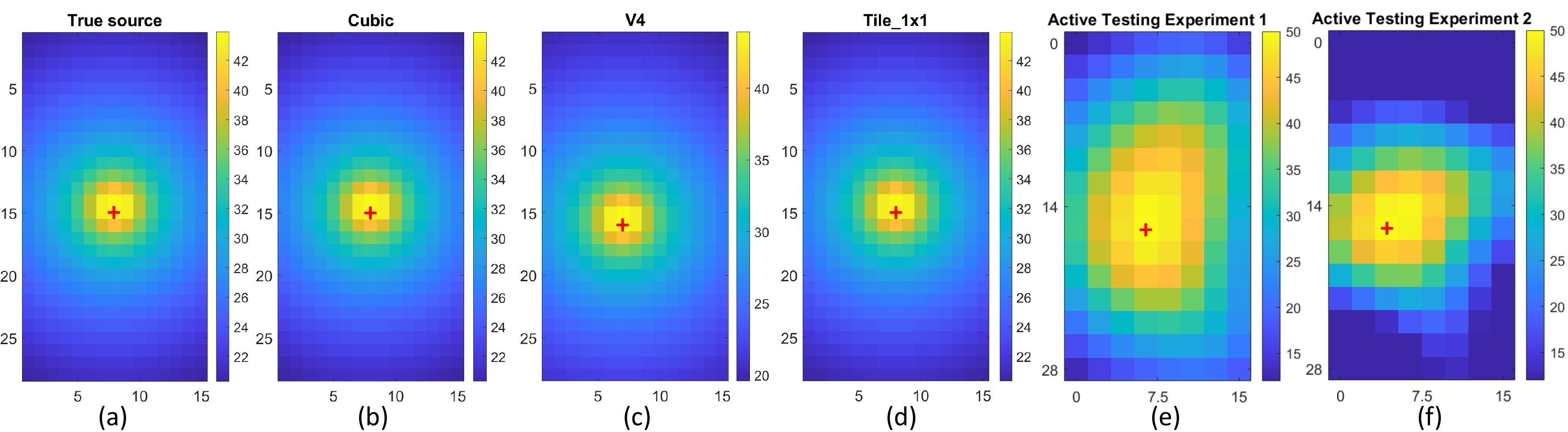}
\caption{The heat map (a) corresponds to the true source. Images (b) to (d) shows the estimated source using Cubic, V4, and Tile\_1x techniques. Images (e) and (f) correspond to two active sensing experiments. Estimated source locations are shown as a red +.}
\label{fig:source_locations}
\end{figure*}

We conducted eight tests to collect RSSI data using two different maps, as shown in Fig.~\ref{fig:maps}, and calculated the errors between the true source location and the estimated source. We analyzed the collected data using three techniques, including our proposed method. The box and whisker plots of the three analyzed techniques are shown in Fig.~\ref{fig:boxplot}. The three techniques are as follows.
\begin{itemize}
    \item \textbf{Interpolation technique (Cubic and V4)} We applied cubic and biharmonic spline interpolation methods to find the global maximums of the heat maps. The biharmonic spline method showed smoother interpolation and an overall lower error compared to the cubic method.
    \item \textbf{Peak RSSI} The distance between the peak RSSI and the source was also included in the analysis.
    \item \textbf{Assigning RSSIs to tiles}  Assigning RSSIs to tiles of four different sizes indicated that high overall accuracy correlates with small tile areas. The analysis showed that three methods approximate the source location with relatively equal accuracy. However, our proposed method of assigning RSSIs for tiles of $1\times1$ m area showed better overall performance in terms of accuracy and outliers. Its overall error is around 1 m which is the lowest among the analysed counterparts.
\end{itemize}

The mean distance error and the error convergence graphs are shown in Fig.~\ref{fig:dist_time}. The shortest mean distance between the estimated source and the true source was reported by the proposed method for the $1\times1$ tile size. The distance error increases with the tile size. Here, we manually pick the best performing tile size for further analysis. The tested interpolation methods can generate heat maps from sparse readings and their source localization accuracy is acceptable. Our analysis showed that the peak power reading is a reliable indication of the source location. However, this depends on the environment. For example, in the presence of strong multi-path reflections, superimposed signals can cause false peak power indications.

The error convergence graph of the proposed method for $1\times1$ m tiles is presented in Fig.~\ref{fig:error_time1}. All tests converged to their minimum error values within three minutes. Our technique depends on the accuracy of the coverage path planning algorithm. For the experiments reported here, we set a low time budget for the path planning to generate a sparse map. The source estimation accuracy can be further increased by using a high time budget which allows the robot to traverse to more areas of the map.

The heat maps correspond to the ground truth source location, and the estimated locations from the discussed techniques are shown in Fig.~\ref{fig:source_locations}.

\subsubsection{Comparison with active sensing-based source localization}


A gas source localization was also experimented with using the active sensing technique. 
The radio field spreads out to the surrounding area of the transmitter by following the Friis propagation model. Active sensing is a popular technique used to address such scenarios (radio or gas source finding). In active sensing, the robot is moved to random locations on the map, and the collected readings are used to update a probabilistic map indicating the source location~\cite{schlotfeldt2018anytime,cao2013multi}. The robot normally operates in a known environment but does not follow predefined trajectories. It can take a long time to collect enough measurements for an accurate estimate.

We compare the source localization results from two active sensing experiments (see Fig.\ref{fig:as_err}) with our proposed technique \textit{Tile\_1x} (see Fig.\ref{fig:error_time1}). The same hardware setting was used for both sets of experiments. All 8 tests of \textit{Tile\_1x} converged before 150 seconds, whereas the best active sensing test took more than 2000 seconds to converge. Active sensing can achieve higher accuracy if more time is given. However, our analysis shows that for a critical mission such as gas source localization, time can be a critical factor. The proposed method can reach a reasonable accuracy in source localization much faster than traditional techniques.



\section{CONCLUSIONS}

In this study, we experimented with different radio signal power strengths received by an antenna mounted on a mobile robot. A robot trajectory was generated using a coverage path planning algorithm. The sparse RSSI readings collected along the traverse were used to estimate the radio source location. Our RSSI-based source localization technique is a simple, fast yet effective solution to the standard computation-heavy methods.

This study could lead to new avenues for using mobile robots to explore unknown areas to detect radio sources or other sources with similar characteristics, such as gas leaks. For our future work, we are expanding our investigations to a broader field area with multiple robots. Another potential extension is to apply machine learning techniques to enhance the estimation accuracy under highly uncertain measurements.

\section*{Acknowledgments}
This work was supported by a grant from the Australian Centre for Advanced Defence Research in Robotics and Autonomous Systems, an initiative funded by the Australian Defence Science and Technology Group.

\bibliographystyle{named}
\bibliography{ref}

\end{document}